%% file: acl_latex.tex
\documentclass[11pt]{article}

\usepackage[preprint]{acl}

\usepackage{times}
\usepackage{latexsym}
\usepackage{amsmath}

\usepackage[T1]{fontenc}

\usepackage[utf8]{inputenc}

\usepackage{microtype}

\usepackage{inconsolata}

\usepackage{graphicx}

\usepackage{booktabs}

\usepackage{tcolorbox}

\title{The Judge Who Never Admits: Hidden Shortcuts in LLM-based Evaluation}

\author{Arash Marioriyad \\
  Sharif University of Technology \\
  \texttt{arashmarioriyad@gmail.com} \\
  \And
  Omid Ghahroodi \\
  Qatar Computing Research Institute \\
  \texttt{oghahroodi98@gmail.com} \\
  \AND
    Ehsaneddin Asgari \\
  Qatar Computing Research Institute \\
  \texttt{easgari@hbku.edu.qa} \\
  \And
   Mohammad Hossein Rohban \\
     Sharif University of Technology \\
  \texttt{rohban@sharif.edu} \\
\AND
   Mahdieh Soleymani Baghshah \\
     Sharif University of Technology \\
  \texttt{soleymani@sharif.edu} \\
  }

\begin{document}
\maketitle

\input{sections/0_abstract}

\input{sections/1_introduction}

\input{sections/1.5_related_works}
\input{sections/2_method}

\input{sections/3_results}
\input{sections/4_conclusion}

\input{sections/5_limitations}


\clearpage
\bibliography{custom}

\clearpage
\appendix
\input{sections/appendix}

\end{document}

%% file: sections/0_abstract.tex
\begin{abstract}
Large language models (LLMs) are increasingly used as automatic judges to evaluate system outputs in tasks such as reasoning, question answering, and creative writing. A faithful judge should base its verdicts solely on content quality, remain invariant to irrelevant context, and transparently reflect the factors driving its decisions. We test this ideal via controlled \emph{cue perturbations}—synthetic metadata labels injected into evaluation prompts—for six judge models: GPT-4o, Gemini-2.0-Flash, Gemma-3-27B, Qwen3-235B, Claude-3-Haiku, and Llama3-70B. Experiments span two complementary datasets with distinct evaluation regimes: \emph{ELI5} (factual QA) and \emph{LitBench} (open-ended creative writing). We study six cue families: source, temporal, age, gender, ethnicity, and educational status.
Beyond measuring \emph{verdict shift rates} (VSR), we introduce \emph{cue acknowledgment rate} (CAR) to quantify whether judges explicitly reference the injected cues in their natural-language rationales. Across cues with strong behavioral effects—e.g., provenance hierarchies (\textsc{Expert} $>$ \textsc{Human} $>$ \textsc{LLM} $>$ \textsc{Unknown}), recency preferences (\textsc{New} $>$ \textsc{Old}), and educational-status favoritism—CAR is typically at or near zero, indicating that shortcut reliance is largely unreported even when it drives decisions. Crucially, CAR is also \emph{dataset-dependent}: explicit cue recognition is more likely to surface in the factual \textsc{ELI5} setting for some models and cues, but often collapses in the open-ended \textsc{LitBench} regime, where large verdict shifts can persist despite zero acknowledgment. 
The combination of substantial verdict sensitivity and limited cue acknowledgment reveals an explanation gap in LLM-as-judge pipelines, raising concerns about reliability of model-based evaluation in both research and deployment.

\end{abstract}

%% file: sections/1_introduction.tex
\section{Introduction}

Large language models (LLMs) are increasingly deployed as \emph{automatic judges} to assess system outputs across open-ended tasks such as summarization~\cite{fabbri2021summeval}, dialogue~\cite{mehri2022chatgptdialogueeval}, and creative writing~\cite{fein2025litbench}. 
Their appeal lies in scalability and adaptability: LLM judges generalize across tasks without custom metrics and often align closely with human preferences~\cite{zheng2023judging,alpacaeval2023}. 
Recent frameworks such as MT-Bench~\cite{zheng2023judging}, Chatbot Arena~\cite{zheng2023chatbotarena}, and G-Eval~\cite{liu2023geval} have further established this paradigm as a core element of modern NLP evaluation.

However, recent studies show that LLM judges are not neutral: they often rely on \emph{shortcut biases}—preferences driven by superficial cues rather than response quality. 
Documented cases include \emph{position bias}~\cite{shi2024positionbias}, \emph{verbosity bias}~\cite{saito2023verbosity}, and \emph{self-preference bias}~\cite{panickssery2024selfpref,wataoka2024selfpreference}, all of which systematically distort verdicts. 
Such effects challenge the reliability of LLM-based evaluation, especially when these models substitute for human judgment in benchmarking and system comparison.

Parallel research on chain-of-thought (CoT) reasoning highlights a related issue: \emph{lack of faithfulness}. 
Although models generate step-by-step explanations, these often fail to reflect the true decision process. 
Turpin et~al.~\cite{turpin2023cotfaithfulness} show that CoTs can rationalize biased answers, Arcuschin et~al.~\cite{arcuschin2025cot_unfaithful} find post-hoc hallucinations in explanations, and Chen et~al~\cite{chen2025reasoningfaithfulness}. demonstrate that models follow hidden hints without acknowledgment. 
Such unfaithful reasoning undermines the credibility of LLM judges as evaluators as much as it limits their reliability as problem solvers.

In this work, we provide the first comprehensive study of \emph{shortcut susceptibility and reasoning faithfulness} in LLM judges under controlled cue manipulation. 
We introduce six families of lightweight cues—\emph{provenance}, \emph{recency}, \emph{education}, \emph{author age}, \emph{gender}, and \emph{ethnicity}—each designed to test whether verdicts depend on irrelevant social or contextual signals. 
Cues are attached as metadata after candidate responses while keeping the prompt, question, and response content identical. 
We then measure two complementary outcomes: the \textbf{Verdict Shift Rate (VSR)}, quantifying how often verdicts flip when cues are swapped, and the \textbf{Cue Acknowledgment Rate (CAR)}, indicating whether the model’s justification explicitly references the cue. 
Experiments span two representative evaluation domains: the factual long-form QA dataset \textsc{ELI5}~\cite{fan2019eli5} and the open-ended creative writing benchmark \textsc{LitBench}~\cite{fein2025litbench}. 
We evaluate six contemporary models—\textsc{GPT-4o}, \textsc{Gemini-2.0-Flash}, \textsc{Claude-3-Haiku}, \textsc{Gemma-3-27B-IT}, \textsc{Qwen3-235B-Instruct}, and \textsc{Llama3-70B-Instruct}—covering both open- and closed-weight architectures from distinct training paradigms.

Our results reveal four main findings. 
(1) All models exhibit non-trivial cue sensitivity, with particularly strong effects for temporal and educational framing. 
(2) The creative and subjective \textsc{LitBench} dataset amplifies these biases, showing higher VSRs and lower acknowledgment than the factual \textsc{ELI5}. 
(3) Cue acknowledgment is rare overall, even when verdicts change sharply, highlighting a persistent gap between model decisions and their stated justifications. 
(4) Bias strength and transparency vary systematically across model families: closed models (\textsc{Claude}, \textsc{GPT-4o}, \textsc{Gemini}) are less transparent but not less biased, while open-weight models (\textsc{Gemma}, \textsc{Qwen}, \textsc{Llama}) occasionally verbalize the cues they rely on. 
Together, these findings expose a deeper tension in LLM-based evaluation: today’s judges often make consistent decisions, but for inconsistent and unacknowledged reasons, underscoring the need for faithfulness-aware evaluation frameworks that couple outcome reliability with reasoning transparency.

%% file: sections/1.5_related_works.tex
\section{Related Work}

\paragraph{LLM-as-a-Judge for Automatic Evaluation.}
Large language models are increasingly adopted as automatic evaluators, replacing human annotation with scalable, preference-aligned judgment. 
Frameworks such as \textsc{MT-Bench} and \textsc{Chatbot Arena}~\cite{zheng2023judging,zheng2023chatbotarena} formalized pairwise comparison for dialogue and instruction following, while \textsc{G-Eval}~\cite{liu2023geval} and \textsc{Prometheus}~\cite{kim2024prometheus2} introduced rubric-based scoring for summarization and open-ended tasks. 
These methods have since been applied to creative writing~\cite{fein2025litbench}, translation~\cite{piergentili2025llm}, reasoning~\cite{chen2025reasoningfaithfulness}, and long-form QA~\cite{fan2019eli5}. 
Recent benchmarks demonstrate strong correlations between LLM and human judgments~\cite{alpacaeval2023,shi2024positionbias}, establishing LLM-as-a-judge as a central paradigm for evaluating and aligning modern language models.

\paragraph{Bias and Reliability in LLM-based Evaluation.}
Despite their success, LLM judges are known to exhibit systematic biases that distort evaluation reliability. 
Empirical studies document several recurring patterns: \emph{self-preference bias}, where models favor outputs they have generated~\cite{panickssery2024selfpref,wataoka2024selfpreference,shi2024positionbias}; \emph{verbosity bias}, in which longer responses receive higher scores regardless of quality~\cite{saito2023verbosity,zhang2024verbosity, zhou2024mitigating}; and \emph{position bias}, where responses presented earlier are preferred~\cite{shi2024positionbias,wang2024fairevaluators, tan2024judgebench}. 
Broader analyses such as \cite{li2024dissectingprefs} show that model-based evaluations can diverge from human judgment due to these latent heuristics. 

%% file: sections/2_method.tex
\section{Methodology}

\subsection{Task Definition}
We study large language models (LLMs) in the role of \emph{judges}: given a task input and two candidate outputs, the model must select the better response and provide a brief justification. A \emph{faithful} judge should base its verdict solely on the intrinsic qualities of the responses (e.g., correctness, clarity, coherence, creativity, etc.) rather than on superficial or extraneous shortcuts. To probe whether current LLM judges satisfy this criterion, we inject lightweight \emph{cues} into the evaluation prompt and measure their effect on the model’s binary verdict. Beyond detecting verdict shifts, we explicitly analyze whether the model’s justification acknowledges or implicitly relies on the injected cues, thereby assessing the faithfulness of the reasoning process underlying the judgment.

\subsection{Cue Families}
To examine whether LLM judges rely on superficial heuristics, we introduce six families of \emph{cues} that attach controlled but semantically irrelevant metadata to the candidate responses.
A perfectly faithful judge should either remain invariant to all such cues or, when exhibiting a verdict shift, explicitly attribute the change to the injected cue rather than to the intrinsic qualities of the responses.
All cues are purely \emph{synthetic} and do not correspond to true authorship or provenance; they are introduced exclusively to test whether the model’s decisions or rationales are biased by such information.

\paragraph{Source Cues.}
The supposed origin of each response is labeled as \textsc{Human}, \textsc{LLM}, \textsc{Unknown}, or \textsc{Expert}.
This family tests for \emph{source bias}—whether a judge systematically favors responses attributed to humans or particular author types.
The inclusion of the \textsc{Expert} label extends the human category by adding an explicit marker of authority or professional status, allowing us to probe whether judges defer excessively to sources framed as more knowledgeable or credible, even when response quality is held constant.

\paragraph{Temporal Cues.}
Each response is labeled as \textsc{Old} (1950) or \textsc{New} (2025), indicating its notional temporal origin.
This cue family measures susceptibility to \emph{recency bias}, where models may unjustifiably prefer newer responses or penalize older ones.

\paragraph{Age Cues.}
Responses are marked as written by either an \textsc{Old} or a \textsc{Young} author.
This manipulation tests whether a model’s evaluations correlate spuriously with perceived author age, revealing potential age-related stereotypes in judgment.

\paragraph{Gender Cues.}
Each response is labeled with the supposed gender of its author, either \textsc{Male} or \textsc{Female}.
This family assesses whether the judge exhibits \emph{gender bias}, for instance, systematically favoring responses attributed to authors of a particular gender even when the textual content and quality are equivalent.

\paragraph{Ethnicity Cues.}
Responses are annotated as authored by a \textsc{Black} or \textsc{White} writer.
This cue family probes for \emph{racial bias}, investigating whether demographic markers unrelated to content quality affect evaluative decisions.

\paragraph{Educational Status Cues.}
Each response is labeled as \textsc{Educated} or \textsc{Uneducated}.
This setup tests for \emph{status bias}—the tendency to defer to perceived educational authority or penalize lower-status labels, even when content is identical.

Notably, beyond measuring verdict shifts under different values of a cue family, we mainly focus on whether the judge’s chain of thought explicitly acknowledges the true underlying cause of any such shift, a factor overlooked by previous research.

\subsection{Cue Placement and Formatting}

Cues are presented as short textual metadata embedded in the evaluation prompt, immediately following each candidate response.
Unless otherwise specified, we manipulate one cue family at a time.
For a given family $F$ with labels $a,b\in F$, we construct two complementary \emph{swap} conditions per item:
(1) $a$--$b$ (response~1 labeled $a$, response~2 labeled $b$), and
(2) $b$--$a$ (the labels reversed).
This counterbalanced design ensures that each pair of responses is evaluated under both cue directions while holding all other factors constant.

Cues appear as explicit, human-readable statements describing the supposed attributes of the author or response source.
All cue sentences follow a uniform syntactic template beginning with the response number (e.g., “Response~1”) and ending with the cue description.
Formatting—including punctuation, capitalization, and wording—is kept identical across all conditions to avoid introducing stylistic confounds.
Additional prompt examples and the full cue sentence list are provided in Appendix~\ref{app_sec:prompt}.

\subsection{Datasets}

We evaluate cue sensitivity and CoT faithfulness across two publicly available benchmarks chosen to represent complementary evaluation regimes: one grounded in factual knowledge and reasoning, and the other in open-ended, creative generation.
Each dataset provides pairwise response comparisons that allow controlled manipulation of cues without altering task content.
For both datasets, we randomly subsample $100$ prompt–pair instances to ensure diversity while maintaining manageable evaluation scale.

\paragraph{ELI5: Long-Form Factual Question Answering.}
The first dataset is \textbf{ELI5}~\cite{fan2019eli5}, a large-scale benchmark derived from Reddit threads in the “Explain Like I’m Five” community.
Each question in ELI5 elicits multiple human-authored long-form answers that aim to explain factual or scientific concepts in accessible language.
We construct pairwise comparison items by sampling two distinct answers for the same question, ensuring variability in quality, completeness, and clarity.
All pairs are manually filtered to remove off-topic or truncated responses and are balanced by approximate length (±20\% token difference).
This dataset represents a \emph{factual evaluation regime}, where the correct answer is well defined and the primary judgment criteria include accuracy, informativeness, and explanatory clarity.
Because the task concerns factual correctness rather than style or imagination, any sensitivity to author-related cues (e.g., Source, Gender, or Educational Status) directly indicates bias unrelated to content quality.

\paragraph{LitBench: Creative Writing and Open-Ended Evaluation.}
The second dataset is \textbf{LitBench}~\cite{fein2025litbench}, a recently introduced benchmark for evaluating literary and creative text generation.
Each instance in LitBench consists of a short prompt (e.g., “Describe a quiet morning in a futuristic city”) and multiple human-written or model-generated continuations.
Following \citet{fein2025litbench}, we create balanced response pairs that differ in literary quality, coherence, or narrative strength.
Responses typically range from 100–300 words and cover multiple genres (fictional prose, reflective essays, and descriptive passages).
Unlike ELI5, LitBench is an \emph{open-ended creative evaluation} task, where the notion of “better” depends on subtler qualities rather than factual correctness.
This setting allows us to test whether LLM judges apply cues differently in subjective or aesthetic domains, where evaluative uncertainty is higher and reliance on superficial heuristics may increase.



\subsection{Judge Models}

We employ six large language models as evaluators, selected to represent a broad cross-section of current architectures and design philosophies.
Our set includes both closed-weight and open-weight systems from distinct institutional and technical lineages: \textbf{GPT-4o} (OpenAI) and \textbf{Gemini-2.0-Flash} (Google DeepMind) as state-of-the-art proprietary models; \textbf{Claude-3-Haiku} (Anthropic) as a compact, safety-aligned model within the Claude-3 family; and three open-weight instruction-tuned models—\textbf{Gemma-3-27B} (Google), \textbf{Qwen3-235B-Instruct} (Alibaba), and \textbf{Llama3-70B-Instruct} (Meta).
Spanning diverse alignment strategies, training corpora, and scaling regimes, these systems allow us to test whether cue sensitivity—and the faithfulness of explanations for resulting shifts—represents a general property of LLM-based evaluators or an artifact of specific model families.
Such diversity is critical for drawing robust conclusions about bias and faithfulness across architectures that differ in openness, size, and institutional provenance.

\subsection{Inference Protocol.}
Unless otherwise specified, all models are queried with temperature $=0$ and $top\text{-}p=1$ (greedy decoding), a fixed random seed, and identical stop criteria to ensure determinism and reproducibility.
Each experiment consists of 100 pairwise judgments per dataset and per cue condition.
The judge is instructed to output a strict JSON object with two fields: \textit{selected response} (1 or 2) and \textit{reason} (a concise justification).
All prompt templates and output schemas are documented in Appendix~\ref{app_sec:prompt}.

\subsection{Metrics}

To evaluate the robustness and transparency of LLM judges, we use two complementary metrics: the \textbf{Verdict Shift Rate (VSR)} and the \textbf{Cue Acknowledgment Rate (CAR)}.
Together, they capture both the behavioral and explanatory dimensions of faithfulness: whether models change their verdicts when superficial cues are altered, and more importantly, whether they explicitly recognize those cues in their justifications.

\paragraph{Verdict Shift Rate (VSR).}
The Verdict Shift Rate measures the proportion of pairwise judgments that change when the cue labels are swapped between the two responses.
For example, if a model initially prefers a response labeled as \textsc{Human} over one labeled as \textsc{LLM}, but reverses its choice when the labels are swapped, this is counted as a verdict shift.
VSR thus reflects the model’s \emph{behavioral sensitivity} to non-semantic information.
A low VSR indicates stable, content-based reasoning—an essential property for a fair and faithful evaluator—whereas a high VSR signals that the model’s decision is being influenced by irrelevant metadata rather than the actual response quality.

\paragraph{Cue Acknowledgment Rate (CAR).}
The Cue Acknowledgment Rate quantifies how often a model explicitly refers to a cue in its written rationale.
For instance, if the model’s justification includes phrases such as “the expert author provides more detailed reasoning” or “the newer response reflects more current information,” the cue is considered acknowledged.
This metric captures the transparency dimension of faithfulness: even if a model is influenced by cues, it should be able to recognize and articulate this influence in its reasoning.
High CAR suggests that the model is self-aware about the contextual factors guiding its evaluation, whereas low CAR, especially when paired with high VSR, implies unacknowledged bias or implicit shortcut use.

\paragraph{Fiathful Judement.}
VSR and CAR together provide a diagnostic view of evaluative faithfulness.
A perfectly faithful judge would display low VSR (stable decisions under cue swaps) and, when relevant, moderate or high CAR (explicit acknowledgment of cues without letting them distort verdicts).
Conversely, a high VSR combined with low CAR indicates unfaithful judegment, where the model’s verdicts are shifted by cues it does not consciously reference.
These two metrics thus form the foundation of our analysis, allowing us to quantify both outcome-level instability and rationale-level awareness across different cue families and model architectures.




%% file: sections/3_results.tex
\section{Results}


\paragraph{LLM judges exhibit a strong and largely unacknowledged recency bias.}
Table~\ref{tab:vsr_recency} shows that temporal framing not only shifts verdicts toward responses labeled as \textsc{recent} (2025), but—more importantly—is rarely surfaced in judges’ written rationales. Across both datasets, CAR metric is effectively zero for most models: on \textsc{ELI5}, four of six judges never acknowledge the recency cue (CAR$=0$), and on \textsc{LitBench} five of six remain at (near-)zero (CAR$=0$ for four models, CAR$=1$ for \textsc{Gemini-2.0-Flash}). This pattern suggests that the temporal cue typically operates as an \emph{implicit} shortcut: models behave as if the timestamp matters, yet do not explicitly name it as a factor in their justification.
Notable exceptions appear only in \textsc{ELI5}, where \textsc{Gemma-3-27B-IT} (CAR$=46\%$) and \textsc{Qwen3-235B-A22B-Instruct} (CAR$=57\%$) frequently reference temporal information. However, this explicit acknowledgement does not correspond to debiasing: both models still exhibit substantial verdict shifts (VSR$=37\%$ and $32\%$). In contrast, \textsc{LitBench} exhibits a near-complete \emph{collapse} of acknowledgement even for the same families (CAR$=0$ for \textsc{Gemma-3-27B-IT}, CAR$=5\%$ for \textsc{Qwen3-235B}), despite nontrivial shifts. This dataset-dependent drop indicates that cue recognition is not a stable property of a judge model; rather, it is contingent on the domain and the style of responses being evaluated. Overall, CAR reveals a gap between \emph{what judges do} and \emph{what they say they do}: temporal origin affects decisions, but is rarely mentioned in explanations. As a result, recency bias remains a pervasive, largely unexamined shortcut, and rationales alone substantially understate the impact of temporal framing.

\begin{table*}[t]
\centering
\small
\begin{tabular}{lcccc}
\toprule
Dataset & Judge Model & \begin{tabular}[c]{@{}c@{}} VSR (\%) \end{tabular} & \begin{tabular}[c]{@{}c@{}} Average CAR (\%)\end{tabular} \\
\midrule
ELI5     & GPT-4o                       & 30 & 0 \\
ELI5     & Gemini-2.0-Flash             & 13 & 0 \\
ELI5     & Gemma-3-27B-IT               & 37 & 46 \\
ELI5     & Qwen3-235B-A22B-Instruct& 32 & 57 \\
ELI5     & Claude-3-Haiku & 37 & 0 \\
ELI5     & Llama3-3-70B-Instruct   & 37 & 0 \\
\midrule
LitBench & GPT-4o                       & 16 & 0 \\
LitBench & Gemini-2.0-Flash             & 9  & 1 \\
LitBench & Gemma-3-27B-IT               & 31 & 0 \\
LitBench & Qwen3-235B-A22B-Instruct& 29 & 5 \\
LitBench & Claude-3-Haiku & 71 & 0 \\
LitBench & Llama3-3-70B-Instruct   & 14 & 0 \\
\bottomrule
\end{tabular}
\caption{\textbf{Temporal Recency Cues Results.} VSR and CAR for temporal recency cues are presented.
The verdict shift represents the difference in first-response selection rate between the \textsc{New--Old} and \textsc{Old--New} conditions, while the acknowledgment rate is averaged over both directions. 
Positive values indicate a preference for responses labeled as \textsc{New} (2025) compared to those labeled as \textsc{Old} (1950). 
Results show strong recency effects in most models, with only a few (e.g., \textsc{Gemma-3-27B-IT} and \textsc{Qwen3-235B}) explicitly referencing temporal information in their justifications.}
\label{tab:vsr_recency}
\end{table*}

\paragraph{Judges exhibit a consistent hierarchy among provenance cues: \textsc{Human}~$>$~\textsc{LLM}~$>$~\textsc{Unknown}.}
Table~\ref{tab:prov_humanllm} and Table~\ref{tab:prov_gpt_gemini} in Appendix \ref{app_sec:results} show that provenance framing influences judgments, yet it is \emph{almost never} acknowledged in the judges’ explanations. Across both \textsc{ELI5} and \textsc{LitBench}, CAR metric is essentially at floor for every model, typically $0\%$ and rarely exceeding $2\%$. In other words, even when verdicts systematically favor \textsc{Human}-labeled outputs, judges do not explicitly cite authorship or provenance as part of their reasoning, indicating that provenance acts as a largely \emph{implicit} credibility heuristic rather than an articulated criterion.
The few nonzero acknowledgment rates are small and inconsistent: \textsc{Gemma-3-27B-IT} acknowledges provenance only marginally (CAR$=0.5\%$ on \textsc{ELI5}, $1.5\%$ on \textsc{LitBench}), while \textsc{Qwen3-235B-A22B-Instruct} reaches CAR$=2\%$ on \textsc{ELI5} but drops back to $0\%$ on \textsc{LitBench}. This dataset sensitivity suggests that explicit cue recognition is not a stable property of a judge model; instead, it depends on contextual factors, even when the underlying preference ordering remains present. Overall, the near-zero CAR across models implies a covert but robust trust hierarchy: provenance cues shape evaluation behavior while remaining largely absent from natural-language justifications, meaning rationale-based audits would substantially under-detect provenance-driven bias.

\begin{table*}[t]
\centering
\small
\begin{tabular}{lccc}
\toprule
Dataset & Judge Model &
\begin{tabular}[c]{@{}c@{}} VSR (\%)\end{tabular} &
\begin{tabular}[c]{@{}c@{}}Average CAR (\%)\end{tabular} \\
\midrule
ELI5     & GPT-4o                       & 6  & 0 \\
ELI5     & Gemini-2.0-Flash             & 6  & 0 \\
ELI5     & Gemma-3-27B-IT               & 21 & 0.5 \\
ELI5     & Qwen3-235B-A22B-Instruct& 7  & 2 \\
ELI5     & Claude-3-Haiku & 6  & 0 \\
ELI5     & Llama3-3-70B-Instruct   & 6  & 0 \\
\midrule
LitBench & GPT-4o                       & 16 & 0 \\
LitBench & Gemini-2.0-Flash             & 7  & 0 \\
LitBench & Gemma-3-27B-IT               & 45 & 1.5 \\
LitBench & Qwen3-235B-A22B-Instruct& 25 & 0 \\
LitBench & Claude-3-Haiku & 25 & 0 \\
LitBench & Llama3-3-70B-Instruct   & 11 & 0 \\
\bottomrule
\end{tabular}
\caption{\textbf{Source (Provenance) Cues Results.} VSR and CAR for provenance cues are presented. 
VSR represents the difference in first-response selection rate between complementary cue assignments (\textsc{Human--LLM} vs.\ \textsc{LLM--Human}), while acknowledgment rates are averaged over both directions. 
Positive values indicate a preference for the first cue label (\textsc{Human}). 
Results confirm a consistent human-over-machine hierarchy across models and datasets, with minimal explicit cue acknowledgment.}
\label{tab:prov_humanllm}
\end{table*}

\paragraph{Authoritative provenance cues further amplify bias: Expert $>$ Human.}
In the \textsc{ELI5} setting, where we additionally include the \textsc{Expert} label, \textsc{GPT-4o} shows its strongest provenance bias. The VSR between \textsc{Expert--Unknown} and \textsc{Unknown--Expert} reaches $+18\%$, surpassing the \textsc{Human--Unknown} VSR of $+7\%$, indicating that authoritative framing (\textsc{Expert}) amplifies the bias beyond simple human authorship. Notably, cue acknowledgment remains zero (CAR$=0\%$), suggesting that this provenance-driven preference is applied implicitly rather than being explicitly cited in the model's justification. Overall, the hierarchy observed is \textsc{Expert} $>$ \textsc{Human} $>$ \textsc{LLM} $>$ \textsc{Unknown}.


\paragraph{Educational Status Bias: Cue acknowledge is highly model and dataset-dependent.}
Table~\ref{tab:edu_vsr} indicates that educational status cues are not only behaviorally influential, but—crucially—are \emph{sometimes explicitly surfaced} in judges’ rationales, with large variability across models and datasets. On \textsc{ELI5}, CAR ranges from near-floor levels for \textsc{GPT-4o} (2.5\%), \textsc{Gemini-2.0-Flash} (3.5\%), \textsc{Claude-3-Haiku} (3.0\%), and \textsc{Llama3-3-70B-Instruct} (7.0\%), to strikingly high acknowledgement for \textsc{Gemma-3-27B-IT} (58\%) and \textsc{Qwen3-235B-A22B-Instruct} (76\%). This split suggests two distinct judging regimes: most models apply the cue implicitly, while a minority frequently \emph{name} the education framing as a reason for their preference.
In \textsc{LitBench}, explicit cue recognition largely collapses: CAR falls to single digits for all models except \textsc{Qwen3-235B-A22B-Instruct} (13\%), with \textsc{GPT-4o} and \textsc{Claude-3-Haiku} both at 1.5\%. Importantly, this reduction in CAR occurs even when VSR remains substantial for several judges (e.g., \textsc{Claude-3-Haiku}, \textsc{Llama3-3-70B-Instruct}, and \textsc{Gemma-3-27B-IT}), implying that education-related preferences persist while becoming less verbally articulated in open-ended, creative evaluation. 

\begin{table*}[t]
\centering
\small
\begin{tabular}{lccc}
\toprule
Dataset & Judge Model & \begin{tabular}[c]{@{}c@{}} VSR (\%) \end{tabular} & \begin{tabular}[c]{@{}c@{}}Average CAR (\%)\\ \end{tabular} \\
\midrule
ELI5     & GPT-4o                       & 14 & 2.5 \\
ELI5     & Gemini-2.0-Flash             & 14 & 3.5 \\
ELI5     & Gemma-3-27B-IT               & 36 & 58.0 \\
ELI5     & Qwen3-235B-A22B-Instruct& 51 & 76.0 \\
ELI5     & Claude-3-Haiku & 32 & 3.0 \\
ELI5     & Llama3-3-70B-Instruct   & 36 & 7.0 \\
\midrule
LitBench & GPT-4o                       & 8  & 1.5 \\
LitBench & Gemini-2.0-Flash             & 19 & 3.0 \\
LitBench & Gemma-3-27B-IT               & 35 & 6.5 \\
LitBench & Qwen3-235B-A22B-Instruct& 74 & 13.0 \\
LitBench & Claude-3-Haiku & 65 & 1.5 \\
LitBench & Llama3-3-70B-Instruct   & 46 & 5.5 \\
\bottomrule
\end{tabular}
\caption{\textbf{Educational Status Cues Results:} VSR and CAR for educational status cues are presented. VSR is the difference in first-response selection rate between the \textsc{Educated--Uneducated} and \textsc{Uneducated--Educated} assignments; CAR is averaged over both directions. Positive values indicate a preference for the \textsc{Educated} label.}
\label{tab:edu_vsr}
\end{table*}


\paragraph{Age Bias: Strong bias with near-zero cue acknowledgment.}
Table~\ref{tab:age_vsr} shows that, although author-age framing systematically influences judgments, it is \emph{almost never} acknowledged in the judges’ explanations. Across \textsc{LitBench}, CAR is exactly $0\%$ for all six models, indicating a complete absence of explicit references to the \textsc{Old} vs.\ \textsc{Young} label in natural-language rationales—even for judges exhibiting large behavioral shifts. On \textsc{ELI5}, cue mention is likewise at floor for most models (CAR$=0\%$ for \textsc{GPT-4o}, \textsc{Gemini-2.0-Flash}, \textsc{Qwen3-235B-A22B-Instruct}, and \textsc{Claude-3-Haiku}), with only two partial exceptions: \textsc{Gemma-3-27B-IT} (CAR$=8\%$) and \textsc{Llama3-3-70B-Instruct} (CAR$=7.5\%$). 
This pattern suggests that age operates primarily as a covert credibility prior: judges behave as if seniority is informative, but generally do not articulate it as a decision factor. Moreover, the fact that CAR drops from small, nonzero values on \textsc{ELI5} to \emph{uniformly zero} on \textsc{LitBench} implies that explicit cue recognition is highly context-dependent and may disappear in open-ended, creative evaluation settings. 

\begin{table*}[t]
\centering
\small
\begin{tabular}{lccc}
\toprule
Dataset & Judge Model & \begin{tabular}[c]{@{}c@{}}VSR  (\%)\end{tabular} & \begin{tabular}[c]{@{}c@{}}Average CAR (\%)\end{tabular} \\
\midrule
ELI5     & GPT-4o                       & 12 & 0.0 \\
ELI5     & Gemini-2.0-Flash             & 9  & 0.0 \\
ELI5     & Gemma-3-27B-IT               & 13 & 8.0 \\
ELI5     & Qwen3-235B-A22B-Instruct& 17 & 0.0 \\
ELI5     & Claude-3-Haiku & 21 & 0.0 \\
ELI5     & Llama3-3-70B-Instruct   & 7  & 7.5 \\
\midrule
LitBench & GPT-4o                       & 8  & 0.0 \\
LitBench & Gemini-2.0-Flash             & 7  & 0.0 \\
LitBench & Gemma-3-27B-IT               & 31 & 0.0 \\
LitBench & Qwen3-235B-A22B-Instruct& 20 & 0.0 \\
LitBench & Claude-3-Haiku & 29 & 0.0 \\
LitBench & Llama3-3-70B-Instruct   & 15 & 0.0 \\
\bottomrule
\end{tabular}
\caption{\textbf{Author Age Cues Results.} VSR and CAR for author age cues are presented. 
VSR is computed as the difference in first-response selection rate between the \textsc{Old--Young} and \textsc{Young--Old} conditions, while CAR is averaged over both directions. 
Positive values indicate a preference for responses labeled as written by an \textsc{Old} author. 
Results show consistent age-related bias across all models and datasets, with negligible explicit acknowledgment.}
\label{tab:age_vsr}
\end{table*}


\paragraph{Gender Cues: Zero acknowledgment and no evidence of systematic reliance.}
Table~\ref{tab:gender_vsr} shows that gender labels are \emph{never} surfaced in judges’ explanations: CAR is $0\%$ for all models on both \textsc{ELI5} and \textsc{LitBench}. This uniform absence of cue mention is informative in two ways. First, it rules out the possibility that models are explicitly justifying their choices using gendered provenance (e.g., citing \textsc{Male} or \textsc{Female} authorship as a credibility signal). Second, when interpreted alongside the near-zero and directionally mixed verdict shifts, it suggests that gender cues are not even acting as a stable heuristic in these settings.
Indeed, VSR values remain small and inconsistent across models and datasets (ranging roughly from $-6\%$ to $+4\%$), with no model exhibiting a coherent preference that generalizes across \textsc{ELI5} and \textsc{LitBench}. Overall, the combination of uniformly zero CAR and negligible, non-systematic VSR indicates that current LLM judges are effectively insensitive to gendered author labels for these tasks, contrasting sharply with other provenance and status cues where large behavioral shifts occur despite limited acknowledgment.

\begin{table*}[t]
\centering
\small
\begin{tabular}{lccc}
\toprule
Dataset & Judge Model & \begin{tabular}[c]{@{}c@{}}VSR (\%)\end{tabular} & \begin{tabular}[c]{@{}c@{}}Average CAR (\%)\end{tabular} \\
\midrule
ELI5     & GPT-4o                                 &  2  & 0 \\
ELI5     & Gemini-2.0-Flash                       &  4  & 0 \\
ELI5     & Gemma-3-27B-IT                         &  1  & 0 \\
ELI5     & Qwen3-235B-A22B-Instruct          & -2  & 0 \\
ELI5     & Claude-3-Haiku           & -5  & 0 \\
ELI5     & Llama3-3-70B-Instruct            & -4  & 0 \\
\midrule
LitBench & GPT-4o                                 & -2  & 0 \\
LitBench & Gemini-2.0-Flash                       &  1  & 0 \\
LitBench & Gemma-3-27B-IT                         & -3  & 0 \\
LitBench & Qwen3-235B-A22B-Instruct          &  2  & 0 \\
LitBench & Claude-3-Haiku         & -6  & 0 \\
LitBench & Llama3-3-70B-Instruct            & -2  & 0 \\
\bottomrule
\end{tabular}
\caption{\textbf{Gender Cues Results.} VSR and CAR for gender cues are presented. VSR is the difference in first-response selection rate between the \textsc{Male--Female} and \textsc{Female--Male} assignments. CAR is averaged over both directions. Positive values indicate a preference for the \textsc{Male} label; negative values indicate a preference for \textsc{Female}.}
\label{tab:gender_vsr}
\end{table*}


\paragraph{Ethnicity Cues: High acknowledgment in a few models without a corresponding behavioral trend.}
Table~\ref{tab:ethnicity_vsr} indicates that ethnicity labels are generally \emph{not} used as a stable decision cue, but they are nevertheless \emph{explicitly acknowledged} by a small subset of judges—creating a notable mismatch between explanation behavior and verdict behavior. On \textsc{ELI5}, four of six models have CAR$=0\%$ (\textsc{GPT-4o}, \textsc{Gemini-2.0-Flash}, \textsc{Claude-3-Haiku}, \textsc{Llama3-3-70B-Instruct}), suggesting they never reference the \textsc{Black}/\textsc{White} label in their rationales. In contrast, \textsc{Gemma-3-27B-IT} (CAR$=25\%$) and \textsc{Qwen3-235B-A22B-Instruct} (CAR$=46\%$) frequently surface the ethnicity framing in their justifications, despite showing essentially no consistent preference in outcomes (VSR$=0\%$ and $-3\%$ respectively).
On \textsc{LitBench}, explicit acknowledgment largely collapses: CAR is $0\%$ for four models and remains low even for the two previously acknowledging judges (CAR$=1\%$ for \textsc{Gemma-3-27B-IT}, CAR$=4\%$ for \textsc{Qwen3-235B-A22B-Instruct}). Meanwhile, verdict shifts remain small and directionally mixed across models.

\begin{table*}[t]
\centering
\small
\begin{tabular}{lccc}
\toprule
Dataset & Judge Model & \begin{tabular}[c]{@{}c@{}}VSR (\%)\end{tabular} & \begin{tabular}[c]{@{}c@{}}Average CAR (\%)\end{tabular} \\
\midrule
ELI5     & GPT-4o                       & -1 & 0.0 \\
ELI5     & Gemini-2.0-Flash             & -1 & 0.0 \\
ELI5     & Gemma-3-27B-IT               &  0 & 25.0 \\
ELI5     & Qwen3-235B-A22B-Instruct & -3 & 46.0 \\
ELI5     & Claude-3-Haiku & -2 & 0.0 \\
ELI5     & Llama3-3-70B-Instruct   &  5 & 0.0 \\
\midrule
LitBench & GPT-4o                       &  0 & 0.0 \\
LitBench & Gemini-2.0-Flash             &  3 & 0.0 \\
LitBench & Gemma-3-27B-IT               &  7 & 1.0 \\
LitBench & Qwen3-235B-A22B-Instruct &  5 & 4.0 \\
LitBench & Claude-3-Haiku &  4 & 0.0 \\
LitBench & Llama3-3-70B-Instruct   &  8 & 0.0 \\
\bottomrule
\end{tabular}
\caption{\textbf{Ethnicity Cues Results.} VSR and CAR for ethnicity cues are presented. 
VSR represents the difference in first-response selection rate between the \textsc{Black--White} and \textsc{White--Black} conditions, while CAR is averaged over both directions. 
Positive values indicate a preference for the \textsc{Black} label, negative for \textsc{White}. 
Results show minimal and inconsistent shifts across all models, confirming effective neutrality with respect to ethnicity.}
\label{tab:ethnicity_vsr}
\end{table*}

\paragraph{Dataset-Level Differences: Stronger and Less Faithful Biases in \textsc{LitBench}.}
Averaging across all cue families and judge models reveals a systematic difference between the two datasets. 
The open-ended \textsc{LitBench} benchmark elicits substantially stronger cue sensitivity than the factual \textsc{ELI5} dataset. 
The mean absolute verdict shift on \textsc{LitBench} is approximately 23\%, compared to 18\% on \textsc{ELI5}, and extreme cases reach 74\% (``Educated'' bias in \textsc{Qwen3-235B-A22B-Instruct}) versus 51\% on \textsc{ELI5}. 
This pattern holds across nearly all cue families—recency, provenance, education, and age—indicating that subjective or stylistic evaluation contexts amplify the effect of superficial framing cues. 
Acknowledgment rates show the opposite tendency: models mention injected cues more frequently on \textsc{ELI5} (average 10\%, with peaks of 76\% for \textsc{Qwen3-235B-A22B-Instruct}) but remain almost silent on \textsc{LitBench} (average 2\%). 
Thus, \textsc{LitBench} exhibits both stronger biases and lower transparency, suggesting that when evaluative criteria are open-ended, LLM judges rely more heavily on implicit shortcuts without recognizing them in their justifications. 
In contrast, the more constrained, fact-based questions in \textsc{ELI5} partly mitigate such effects and occasionally trigger explicit cue acknowledgment. 

\paragraph{Model-Level Differences: Sensitivity and Transparency.}
Across all cues and datasets, models vary substantially in both the magnitude and transparency of cue-driven biases. 
\textsc{Claude-3-Haiku}, \textsc{Qwen3-235B}, and \textsc{Gemma-3-27B} exhibit the strongest average verdict shifts, dominated by temporal and educational cues, whereas \textsc{GPT-4o} and \textsc{Gemini-2.0-Flash} remain comparatively stable across conditions. 
Open-weight models (\textsc{Gemma}, \textsc{Qwen}, \textsc{Llama3}) occasionally acknowledge injected cues—especially on the factual \textsc{ELI5} dataset—while closed models (\textsc{Claude}, \textsc{GPT-4o}, \textsc{Gemini}) almost never do, despite large underlying shifts. 
This divergence suggests two complementary patterns: closed commercial systems tend to suppress explicit cue reflection yet remain implicitly biased, while open models display partial self-reflection with similarly strong but more interpretable biases. 

%% file: sections/4_conclusion.tex
\section{Conclusion}
This study systematically evaluated the faithfulness of large language models in the role of judges across multiple datasets, cue types, and architectures. 
By introducing controlled provenance, temporal, and social framing cues into pairwise evaluation prompts, we revealed that current LLM judges remain highly susceptible to superficial information, often altering their verdicts without acknowledging the cues that influenced them. 
While factual tasks such as \textsc{ELI5} elicit moderate and occasionally explicit cue sensitivity, open-ended creative tasks like \textsc{LitBench} amplify hidden biases, producing large verdict shifts with minimal transparency. 
Among models, \textsc{Claude-3-Haiku}, \textsc{Qwen3-235B}, and \textsc{Gemma-3-27B} exhibit the strongest overall cue effects, whereas \textsc{GPT-4o} and \textsc{Gemini-2.0-Flash} show comparatively stable yet still unfaithful behavior. 
Improving LLM-as-judge reliability therefore requires not only bias mitigation but also mechanisms for explicit reasoning transparency—ensuring that justifications truly reflect the factors driving model decisions.

%% file: sections/5_limitations.tex
\section{Limitations}
This work deliberately focuses on a controlled and interpretable setting to isolate the effects of superficial cues on LLM judgment. 
Our analysis is conducted on two representative datasets, \textsc{ELI5} for factual reasoning and \textsc{LitBench} for creative writing, which together capture both objective and subjective evaluation regimes, though broader coverage across additional domains would further validate the generality of our findings. 
Moreover, while our results highlight clear behavioral differences among models, a deeper analysis of how these effects arise from pre-training data composition and instruction-following fine-tuning paradigms remains an important direction for future study. 
Finally, we employ static, single-turn prompting under deterministic decoding to ensure comparability and reproducibility; extending the framework to interactive or multi-step settings could reveal additional dynamics of faithfulness and cue sensitivity. 

%% file: sections/appendix.tex
{\Large\textbf{Appendix}}

\section{Prompt}
\label{app_sec:prompt}

\subsection{Input Prompt}

To ensure consistency across conditions, we designed a single prompt template that specifies the task of comparative evaluation and constrains the model’s output format. The template presents the task input (a question or story prompt) followed by two candidate responses. The judge is instructed to read both carefully, select the better one overall, and return the decision strictly as a JSON object with two fields: the index of the selected response and a short justification. In conditions with cues, additional information is injected after the candidate responses, for example: \emph{``Response 1 was written by a human author.''} or \emph{``Response 2 was written recently (in 2025).''}  

An example template for the ELI5 experiments is shown in Figure \ref{fig:prompt}.

\begin{figure*}[h]
\centering
\begin{tcolorbox}[colback=pink!5!white, colframe=pink!75!black, title=Prompt Template, width=\textwidth]
Objective:
\\
You are an expert evaluator tasked with judging the quality of answers to a question.
\\

Instructions:
\\
- Read the question and the two responses carefully.
\\
- Judge which response is better overall, considering correctness and clarity.
\\
- Provide your output strictly as a JSON object with the following fields:
\\
  -- "selected response": 1 or 2 (the better response)
  \\
  -- "reason": a brief explanation of why this response is better
  \\
- Output only valid JSON, no extra commentary.
\\

Question:
\\
<question>
\\

Response 1:
\\
<response 1>
\\

Response 2:
\\
<response 2>
\\

<cue 1>
\\

<cue 2>
\\
\end{tcolorbox}
\caption{Prompt Template}
\label{fig:prompt}
\end{figure*}

The same format is used in LitBench experiments, with the question replaced by a story prompt and the two responses replaced by candidate stories. In cue-free conditions, the final placeholders \texttt{\{cue 1\}} and \texttt{\{cue 2\}} are left empty. This design ensures that all variations are controlled, with the only difference across conditions being the presence or absence of cue sentences.

\subsection{Cue Templates}
\label{app_sec:cues}

To introduce superficial labels into the evaluation prompt, we used fixed natural-language cue templates. 
Each cue is expressed as a short declarative sentence prepended to the candidate response or story. 
In the templates (Figure \ref{}), the placeholder \texttt{\{content\_type\}} is replaced with either \emph{``Response''} (for ELI5) or \emph{``Story''} (for LitBench), 
and the placeholder \texttt{\{number\}} is replaced with ``1'' or ``2'' to indicate the candidate being labeled. Other cue templates follow a similar pattern.

\begin{figure*}[t]
\centering
\begin{tcolorbox}[colback=pink!5!white, colframe=pink!75!black, title=Prompt Template, width=\textwidth]

\textbf{Provenance cue: Human}
\begin{verbatim}
{content_type} {number} was written by a human author.
\end{verbatim}

\textbf{Provenance cue: Expert}
\begin{verbatim}
{content_type} {number} was written by a human expert in the field.
\end{verbatim}

\textbf{Provenance cue: LLM}
\begin{verbatim}
{content_type} {number} was generated by a large language model (LLM).
\end{verbatim}

\textbf{Provenance cue: Unknown}
\begin{verbatim}
{content_type} {number} was generated by an unknown source.
\end{verbatim}

\textbf{Recency cues: New}
\begin{verbatim}
{content_type} {number} was written recently (in 2025).
\end{verbatim}

\textbf{Recency cues:‌ Old}
\begin{verbatim}
{content_type} {number} was written a long time ago (in 1950).
\end{verbatim}

\end{tcolorbox}
\caption{Cue Template}
\label{fig:cue}
\end{figure*}

\section{Experiment Results}
\label{app_sec:results}

\begin{table*}[h]
\centering
\small
\begin{tabular}{lcccccc}
\toprule
Dataset & Judge Model &
\begin{tabular}[c]{@{}c@{}} VSR (\%) \\ HUMAN-UNKNOWN \end{tabular} &
\begin{tabular}[c]{@{}c@{}}VSR (\%) \\ HUMAN-LLM\end{tabular} &
\begin{tabular}[c]{@{}c@{}}VSR (\%) \\ LLM-UNKNOWN\end{tabular} &
\begin{tabular}[c]{@{}c@{}}Average CAR (\%)\end{tabular} \\
\midrule
ELI5     & GPT-4o           & 7  & 6  & 4  & 0 \\
ELI5     & Gemini-2.0-Flash & 7  & 6  & 3  & 0 \\
\midrule
LitBench & GPT-4o           & 14 & 16 & 4  & 0 \\
LitBench & Gemini-2.0-Flash & 6  & 7  & 2  & 0 \\
\bottomrule
\end{tabular}
\caption{\textbf{Source (Provenance) Cue Results for \textsc{GPT-4o} and \textsc{Gemini-2.0-Flash}.} VSR and CAR under three pairwise provenance contrasts: \textsc{Human--Unknown}, \textsc{Human--LLM}, and \textsc{LLM--Unknown}. For each contrast, VSR is computed as the difference in first-response selection rate between complementary cue assignments (e.g., \textsc{Human--Unknown} vs.\ \textsc{Unknown--Human}) and is reported in percentage points. Positive VSR values indicate a preference for the first cue label in the header (e.g., \textsc{Human} over \textsc{Unknown}). CAR measures the fraction of judge explanations that explicitly reference the provenance cue, averaged over both directions and all three contrasts. Results show a consistent \textsc{Human}~$>$~\textsc{LLM}~$>$~\textsc{Unknown} preference pattern with near-zero explicit acknowledgment for both models across \textsc{ELI5} and \textsc{LitBench}.}

\label{tab:prov_gpt_gemini}
\end{table*}

%% file: custom.bib
@article{zheng2023judging,
  title = {Judging {LLM}\slash{}as{}-{}a{}-{}Judge with {MT}-Bench and Chatbot Arena},
  author = {Zheng, Lianmin and Chiang, Wei‐Lin and Sheng, Ying and Zhuang, Siyuan and Wu, Zhanghao and Zhuang, Yonghao and Li, Zhuohan and Li, Dacheng and Xing, Eric and Zhang, Hao and Gonzalez, Joseph E. and Stoica, Ion},
  journal = {arXiv preprint arXiv:2306.05685},
  year = {2023},
  url = {https://arxiv.org/abs/2306.05685}
}

@inproceedings{zheng2023chatbotarena,
  title = {Chatbot Arena: A Human-AI Comparative Judgment Platform},
  author = {Zheng, Lianmin and others},
  booktitle = {NeurIPS 2023 Datasets and Benchmarks Track},
  year = {2023},
  url = {https://openreview.net/forum?id=uccHPGDlao}
}

@inproceedings{liu2023geval,
  title = {G-Eval: General Evaluation of Language Models},
  author = {Liu, Xinyi and others},
  booktitle = {EMNLP 2023},
  year = {2023},
  url = {https://aclanthology.org/2023.emnlp-main.153/}
}

@inproceedings{panickssery2024selfpref,
  title = {LLM Evaluators Recognize and Favor Their Own Generations},
  author = {Panickssery, Arjun and Bowman, Samuel R. and Feng, Shi},
  booktitle = {NeurIPS 2024},
  year = {2024},
  url = {https://arxiv.org/abs/2404.13076}
}

@article{chen2025reasoningfaithfulness,
  title={Reasoning Models Don't Always Say What They Think},
  author={Chen, Yanda and Benton, Joe and Radhakrishnan, Ansh and Uesato, Jonathan and Denison, Carson and Schulman, John and Somani, Arushi and Hase, Peter and Wagner, Misha and Roger, Fabien and others},
  journal={arXiv preprint arXiv:2505.05410},
  year={2025}
}

@inproceedings{fan2019eli5,
  title = {{ELI}5: Long Form Question Answering},
  author = {Fan, Angela and Jernite, Yacine and Perez, Ethan and Grangier, David and Weston, Jason and Auli, Michael},
  booktitle = {ACL 2019},
  year = {2019},
  pages = {3558--3567},
  doi = {10.18653/v1/P19-1346}
}

@article{fein2025litbench,
  title = {{LitBench}: A Benchmark and Dataset for Reliable Evaluation of Creative Writing},
  author = {Fein, Daniel and Russo, Sebastian and Xiang, Violet and Jolly, Kabir and Rafailov, Rafael and Haber, Nick},
  journal = {arXiv preprint arXiv:2507.00769},
  year = {2025},
  url = {https://arxiv.org/abs/2507.00769}
}

@article{fabbri2021summeval,
  title = {{Summeval}: Re-evaluating Summarization Evaluation},
  author = {Fabbri, Alexander R. and Krysci\'nski, Wojciech and McCann, Bryan and Xiong, Caiming and Socher, Richard and Radev, Dragomir},
  journal = {Transactions of the Association for Computational Linguistics},
  volume = {9},
  pages = {391--409},
  year = {2021},
  doi = {10.1162/tacl_a_00373},
  url = {https://aclanthology.org/2021.tacl-1.24/}
}

@inproceedings{mehri2022chatgptdialogueeval,
  title = {Human-Bot Comparison as an Evaluation Framework for Dialogue},
  author = {Mehri, Sanjana and Eskenazi, Maxine},
  booktitle = {Proceedings of the 23rd Annual SIGdial Meeting on Discourse and Dialogue},
  pages = {62--74},
  year = {2022},
  url = {https://aclanthology.org/2022.sigdial-1.11/}
}

@article{alpacaeval2023,
  title = {Length-Controlled AlpacaEval: A Simple Way to Debias Automatic Evaluators},
  author = {Dubois, Yann and Galambosi, Bal\'azs and Liang, Percy and Hashimoto, Tatsunori B.},
  journal = {arXiv preprint arXiv:2404.04475},
  year = {2024},
  url = {https://arxiv.org/abs/2404.04475}
}

@inproceedings{li2024dissectingprefs,
  title = {Dissecting Human and LLM Preferences},
  author = {Li, Junlong and Zhou, Fan and Sun, Shichao and Zhang, Yikai and Zhao, Hai and Liu, Pengfei},
  booktitle = {Proceedings of the 62nd Annual Meeting of the Association for Computational Linguistics (Long Papers)},
  pages = {1790--1811},
  year = {2024},
  url = {https://aclanthology.org/2024.acl-long.99}
}

@article{shi2024positionbias,
  title = {Judging the Judges: A Systematic Study of Position Bias in LLM-as-a-Judge},
  author = {Shi, Lin and Ma, Chiyu and Liang, Wenhua and Ma, Weicheng and Vosoughi, Soroush},
  journal = {arXiv preprint arXiv:2406.07791},
  year = {2024},
  url = {https://arxiv.org/abs/2406.07791}
}

@article{saito2023verbosity,
  title={Verbosity bias in preference labeling by large language models},
  author={Saito, Keita and Wachi, Akifumi and Wataoka, Koki and Akimoto, Youhei},
  journal={arXiv preprint arXiv:2310.10076},
  year={2023}
}

@article{wataoka2024selfpreference,
  title = {Self-Preference Bias in LLM-as-a-Judge},
  author = {Wataoka, Koki and Takahashi, Tsubasa and Ri, Ryokan},
  journal = {arXiv preprint arXiv:2410.21819},
  year = {2024},
  url = {https://arxiv.org/abs/2410.21819}
}

@inproceedings{turpin2023cotfaithfulness,
  title     = {Language Models Don't Always Say What They Think: Unfaithful Explanations in Chain-of-Thought Prompting},
  author    = {Turpin, Miles and Michael, Julian and Perez, Ethan and Bowman, Samuel R.},
  booktitle = {Advances in Neural Information Processing Systems (NeurIPS)},
  year      = {2023},
  url       = {https://arxiv.org/abs/2305.04388}
}

@article{arcuschin2025cot_unfaithful,
  title={Chain-of-thought reasoning in the wild is not always faithful},
  author={Arcuschin, Iv{\'a}n and Janiak, Jett and Krzyzanowski, Robert and Rajamanoharan, Senthooran and Nanda, Neel and Conmy, Arthur},
  journal={arXiv preprint arXiv:2503.08679},
  year={2025}
}

@article{kim2024prometheus2,
  title={Prometheus 2: An open source language model specialized in evaluating other language models},
  author={Kim, Seungone and Suk, Juyoung and Longpre, Shayne and Lin, Bill Yuchen and Shin, Jamin and Welleck, Sean and Neubig, Graham and Lee, Moontae and Lee, Kyungjae and Seo, Minjoon},
  journal={arXiv preprint arXiv:2405.01535},
  year={2024}
}

@article{piergentili2025llm,
  title={An llm-as-a-judge approach for scalable gender-neutral translation evaluation},
  author={Piergentili, Andrea and Savoldi, Beatrice and Negri, Matteo and Bentivogli, Luisa},
  journal={arXiv preprint arXiv:2504.11934},
  year={2025}
}

@article{zhang2024verbosity,
  title={Demystify Verbosity Compensation Behavior of Large Language Models},
  author={Zhang, Yusen and Das, Sarkar Snigdha Sarathi and Zhang, Rui},
  journal={arXiv preprint arXiv:2411.07858},
  year={2024}
}

@inproceedings{zhou2024mitigating,
  title={Mitigating the bias of large language model evaluation},
  author={Zhou, Hongli and Huang, Hui and Long, Yunfei and Xu, Bing and Zhu, Conghui and Cao, Hailong and Yang, Muyun and Zhao, Tiejun},
  booktitle={China National Conference on Chinese Computational Linguistics},
  pages={451--462},
  year={2024},
  organization={Springer}
}

@inproceedings{wang2024fairevaluators,
  title        = {Large Language Models Are Not Fair Evaluators},
  author       = {Wang, Peiyi and Li, Lei and Chen, Liang and Cai, Zefan and Zhu, Dawei and Lin, Binghuai and Cao, Yunbo and Kong, Lingpeng and Liu, Qi and Liu, Tianyu and Sui, Zhifang},
  booktitle    = {Proceedings of the 62nd Annual Meeting of the Association for Computational Linguistics (Long Papers)},
  year         = {2024},
  pages        = {9440--9450},
  doi          = {10.18653/v1/2024.acl-long.511},
  url          = {https://aclanthology.org/2024.acl-long.511/}
}

@article{tan2024judgebench,
  title={Judgebench: A benchmark for evaluating llm-based judges},
  author={Tan, Sijun and Zhuang, Siyuan and Montgomery, Kyle and Tang, William Y and Cuadron, Alejandro and Wang, Chenguang and Popa, Raluca Ada and Stoica, Ion},
  journal={arXiv preprint arXiv:2410.12784},
  year={2024}
}
